\documentclass[10pt,twocolumn,letterpaper]{article}

\usepackage[accsupp]{axessibility}
\usepackage[pagenumbers]{arxiv}  

\usepackage{graphicx}
\usepackage{wrapfig}
\usepackage{afterpage}
\definecolor{cvprblue}{rgb}{0.21,0.49,0.74}
\usepackage[pagebackref,breaklinks,colorlinks,allcolors=cvprblue]{hyperref}
\usepackage{float} 
\usepackage{graphicx}

\usepackage{bm}
\usepackage{booktabs}

\title{FruitNinja: 3D Object Interior Texture Generation with Gaussian Splatting}

\author{Fangyu Wu \quad Yuhao Chen \\
University of Waterloo
}
\begin{document}
\twocolumn[{%
\renewcommand\twocolumn[1][]{#1}%
\maketitle
\begin{center}
    \centering
    \captionsetup{type=figure}
\includegraphics[width=
0.9\textwidth]{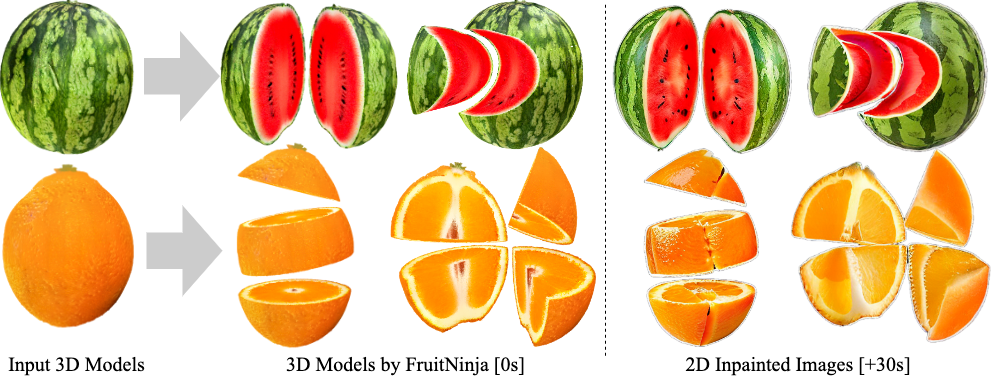}
    \captionof{figure}{FruitNinja generates high-quality interior textures for 3DGS models, enabling real-time view rendering during arbitrary geometric transformations. In contrast, direct 2D inpainting requires additional optimization steps ($\sim$30s), often misaligns edited geometry, and yields inconsistent results per edit.}
\end{center}%
}]
\begin{abstract}
In the real world, objects reveal internal textures when sliced or cut, yet this behavior is not well-studied in 3D generation tasks today. For example, slicing a virtual 3D watermelon should reveal flesh and seeds. Given that no available dataset captures an object's full internal structure and collecting data from all slices is impractical, generative methods become the obvious approach. However, current 3D generation and inpainting methods often focus on visible appearance and overlook internal textures. To bridge this gap, we introduce FruitNinja, the first method to generate internal textures for 3D objects undergoing geometric and topological changes. Our approach produces objects via 3D Gaussian Splatting (3DGS) with both surface and interior textures synthesized, enabling real-time slicing and rendering without additional optimization. FruitNinja leverages a pre-trained diffusion model to progressively inpaint cross-sectional views and applies voxel-grid-based smoothing to achieve cohesive textures throughout the object. Our OpaqueAtom GS strategy overcomes 3DGS limitations by employing densely distributed opaque Gaussians, avoiding biases toward larger particles that destabilize training and  sharp color transitions for fine-grained textures. Experimental results show that FruitNinja substantially outperforms existing approaches, showcasing unmatched visual quality in real-time rendered internal views across arbitrary geometry manipulations.
\end{abstract}
\section{Introduction}
\label{sec:intro}

Generating high-quality interactive 3D objects has diverse applications in fields such as augmented and virtual reality (AR/VR), digital gaming, and advertising. Recent advancements in 3D computer vision, particularly 3D Gaussian Splatting (3DGS) \cite{kerbl2023gaussian}, have introduced efficient techniques for novel view synthesis. Leveraging its explicit point-cloud-based representation and suitability for post-editing tasks, researchers have further explored user-guided editing of 3DGS, including stylization \cite{chen2024gaussianeditor, stylizedgs, liu2023stylegaussian}, deformation\cite{Xie2023PhysGaussianP3, vrgs, borycki2024gaspgaussiansplattingphysicbased, huang2023pointnmoveinteractivescene} , object removal \cite{gaussian_grouping, chen2024gaussianeditor}, inpainting \cite{liu2024infusion, chen2024gaussianeditor} and texture editing \cite{palandra2024gseditefficienttextguidedediting, xu2024texturegs}.  In interactive 3D applications, it's common for users to perform customized or large-scale geometric modifications—such as cutting, tearing, or removing parts of an object's surface, which can reveal internal textures. If the internal Gaussians are not adequately trained, this may expose unrealistic internal structures, thereby compromising overall visual quality. Unfortunately, existing 3DGS editing frameworks primarily focus on manipulating an object's external appearance, while preserving the fidelity of object's internal textures remains under-explored.

Creating novel view synthesis of an object's internal structure, ensuring that the exposed internal texture appears realistic when sliced from any arbitrary angle, is particularly challenging due to scarcity of training data. Current 3D datasets predominantly emphasize the overall geometry and surface texture of objects \cite{shapenet2015, objaverse, uy-scanobjectnn-iccv19, Koch_2019_CVPR}, often lacking details about their interiors. Acquiring data on internal structures typically requires specialized techniques such as X-ray scans, CT imaging, or compiling multiple cross-sectional images. Furthermore, when images are collected by disassembling objects (e.g., cutting objects in half), only partial internal structures are exposed. Reversing these manipulations to access other parts is often impractical, making it difficult to infer the complete internal structure.

Existing methods for texture generation in 3D objects \cite{chen2023text2tex, richardson2023texture, cao2023texfusion} focus on generating surface textures for mesh UV maps, primarily addressing the object's external shell and therefore are not applicable for internal modeling in 3DGS. Some recent studies \cite{vrgs, huang2023pointnmoveinteractivescene} have employed ad-hoc inpainting on newly exposed regions after each editing step to mitigate visual artifacts. However, this per-edit approach can introduce inconsistencies across a series of transformations, and is unsuitable for real-time rendering. PhysGaussian introduced a mechanism designed to enhance the visual quality of 3D models under deformation by internally filling Gaussian particles\cite{Xie2023PhysGaussianP3}. This method discretizes the opacity field onto a 3D grid and uses ray casting to identify and fill void regions based on opacity thresholds, with each filled particle inheriting color and opacity from nearby surface Gaussian kernels. Nonetheless, this approach relies on the unrealistic assumption that internal textures resemble surface textures, while real-world objects often have different internal characteristics, causing inherited properties to fall short of capturing realistic details.

Fortunately, many common objects possess symmetrical features, allowing their cross-sectional views at consistent angles to appear similar. For instance, slicing a watermelon horizontally at different levels consistently reveals similar patterns of skin, flesh, and seeds. Building on this observation, we propose FruitNinja, an effective method for generating 3D internal textures by using only a few cross-sectional views as references. Our method synthesizes the entire interior texture of an object without requiring additional optimization after geometry edits or topology changes. FruitNinja leverages a pre-trained diffusion model to guide the synthesis of cross-sectional views using Score Distillation Sampling (SDS)\cite{poole2022dreamfusion}, thereby jointly training cross-section and surface views. To address inconsistent artifacts from varying generated cross-sectional views, FruitNinja progressively optimizes reference views and applies voxel-grid-based smoothing to seamlessly blend the overall texture. As real-world objects are composed of millions of atoms and molecules, we adopt the OpaqueAtom GS settings inspired by AtomGS \cite{liu2024atomgs}. This overcomes two key limitations of the original  3DGS algorithm: (1) the tendency to optimize larger Gaussians, which limits the density of small Gaussian particles needed for stable training and introduces artifacts during editing; and (2) limitations in the GS ray-marching method, where blending front and back Gaussians compromises modeling abrupt color transitions (e.g., white flesh adjacent to green skin or red flesh in watermelon slices). 

Our contributions are summarized as follows:

\begin{itemize}
    \item We introduce the first method for generating textures for object interiors by progressively inpainting cross-sectional views and applying voxel smoothing, enabling real-time rendering of internal views.
    
    \item We propose an  OpaqueAtom GS strategy for modeling 3D objects with realistic interior textures, allowing arbitrary slicing to reveal fine-grained details while overcoming limitations of 3DGS, including instability from large Gaussians and difficulty with sharp color transitions.
    
    \item We demonstrate that the proposed method effectively synthesizes internal textures across common objects, achieving superior view quality during various geometric transformations compared to existing methods, both qualitatively and quantitatively.
\end{itemize}

\begin{figure*}[t]
    \centering
\includegraphics[width=\textwidth]{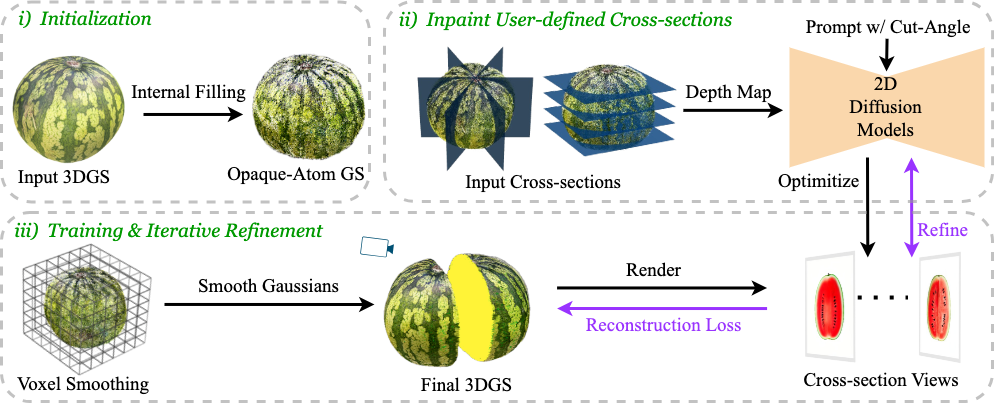}
    \caption{Method Overview. The input 3DGS is first transformed using opaque atomic Gaussians, and void regions within the object are filled with raw particles. For each user-defined cut angle, a reference cross-sectional view is generated via SDS. Subsequently, exterior Gaussians are iteratively masked out to render cross-sectional views, which are jointly trained with randomly selected external views. Each reference cross-sectional view is continuously refined by applying SDS to the existing rendering, while voxel smoothing is applied every $N$ iterations to improve overall consistency.
    }
    \label{fig:placeholder}
\end{figure*}
\section{Related Work}
\label{sec:formatting}

\subsection{3D Representations}
Various 3D representations—such as point clouds \cite{rusu20113d}, meshes \cite{botsch2010polygon}, 3D voxel grids \cite{niessner2013real}, and signed distance functions (SDFs) \cite{park2019deepsdf}—each offers unique advantages for 3D tasks. Recently, Neural Radiance Fields (NeRF) introduced a breakthrough by implicitly encoding 3D scenes through deep networks and volumetric rendering, achieving high-quality reconstructions. Building on this innovation, some research has focused on enabling customized geometry deformations based on NeRF. Methods such as NeRF-Editing \cite{yuan2022nerfediting}, Neural Impostor\cite{liu2023neuralimpostoreditingneural}, and NeuPhysics\cite{qiao2022neuphysics} integrate explicit mesh structures with implicit neural representations to facilitate intuitive and physically accurate manipulations of 3D models. Alternatively, 3D Gaussian Splatting \cite{kerbl2023gaussian} employs an explicit, point-cloud-based approach using Gaussian kernels, offering faster and more easily editable representations. Techniques like GaussianEditor\cite{chen2024gaussianeditor}, GSDeformer\cite{huang2024gsdeformer}, and Mani-GS\cite{gao2024mani} enable user-guided texture and geometry transformations by leveraging Gaussian semantic tracing, cage-based deformation, and mesh adaptation. Additionally, PhysGaussians\cite{Xie2023PhysGaussianP3} and VR-GS\cite{vrgs} have introduced physics-based deformation methods based on 3DGS. AtomGS \cite{liu2024atomgs} enhances 3DGS by introducing Atomized Proliferation to densify small, uniform Gaussians in areas of fine detail, and Geometry-Guided Optimization to align Gaussians with scene geometry, reducing noise and sharpening edges. This approach enables high-fidelity rendering and precise geometry reconstruction, making it ideal for applications that require detailed textures. Inspired by these capabilities, we adopt an OpaqueAtom GS strategy to achieve high fidelity in fine-grained interior textures of 3D models.


\subsection{3D Inpainting}
3D inpainting addresses the challenge of filling missing or masked regions in 3D spaces by generating plausible geometry and textures. Early inpainting works predominantly focus on either geometry completion or texture synthesis, often treating these aspects separately. For instance, methods like \cite{Wang2017ShapeIU, zhang2021unsupervised} concentrate on reconstructing the underlying geometric structures, while others methods like \cite{chen2023text2tex, cao2023texfusion, huang2023nerf_texture, siddiqui2022texturify, richardson2023texture} target the generation of realistic textures for the completed regions. Recent advancements have enabled the simultaneous inpainting of both semantic content and geometric structures, effectively addressing the interplay between these two components. NeRF-based approaches have leveraged features from models like CLIP\cite{clip} to learn and incorporate 3D semantics into the inpainting process, enhancing the contextual relevance and realism of the completed regions\cite{clipnerf2022}. Additionally, several methods have explored inpainting techniques within the framework of 3DGS, benefiting from GS’s rendering efficiency and high-quality reconstruction capabilities \cite{chen2024gaussianeditor, liu2024infusion, huang2023pointnmoveinteractivescene}. However, previous inpainting methods are mainly designed for static scenes, which limits their effectiveness in dynamic settings where untrained interior areas may be exposed. In this paper, we propose a method focused on inpainting internal textures to overcome these limitations.

\section{Method}

In this section, we outline our approach for generating internal textures for 3DGS objects to ensure visual coherence under arbitrary geometric modifications. Our method involves three steps: first, we populate the interior of the 3D object with raw Gaussian particles (Section \ref{sec:gaussian-initialization}). Next, we apply SDS optimization to the input cross-sectional views to produce an initial set of cross-sectional reference views (Section \ref{sec:cross-section-inpaint}). Then, we use these reference views, along with surface views, to jointly train the 3DGS model with OpaqueAtom GS settings (Section \ref{sec:opaque_atom}), iteratively refining the reference views set and applying voxel-based smoothing to enhance texture consistency (Section \ref{sec:refinement}).

\subsection{Preliminary}

\subsubsection{3D Gaussian Splatting} 
\label{sec:gaussian-rendering}

3D Gaussian splatting represents a 3D scene as a set of $N$ Gaussians $\mathcal{G} = \{g_1, g_2, \dots, g_N\}$.
Each Gaussian $g_i$ is characterized by its position $\mathbf{x}_i \in \mathbb{R}^3$, a covariance matrix $\Sigma_i \in \mathbb{R}^{3 \times 3}$ that describes its shape and orientation, color coefficients $\mathbf{c}_i \in \mathbb{R}^k$, and an opacity value $\alpha_i \in \mathbb{R}$.
To render the scene from a new viewpoint, each Gaussian is projected onto the camera plane by applying a viewing transformation, which includes adjusting its covariance matrix $\Sigma_i$ to account for perspective.
For each pixel, the contributions from overlapping Gaussians are blended. 
The influence $\sigma_i$ of Gaussian $g_i$ on a pixel is computed based on its opacity $\alpha_i$ and the Gaussian function evaluated at that pixel.
The final pixel color $C$ is obtained by front-to-back compositing of each Gaussian's contribution:
\begin{equation}
C = \sum_{i=1}^{N} \mathbf{c}_i \, \sigma_i \, \prod_{j=1}^{i-1} \bigl(1 - \sigma_j\bigr)\,.
\label{eqn:gaussian-splatting}
\end{equation}

\subsubsection{Internal Gaussians Initialization}
\label{sec:gaussian-initialization}

Following the internal filling mechanism from PhysGaussian, we fill the 3D object's internal regions with Gaussian primitives.
To identify empty internal regions, we first construct a continuous 3D opacity field $d(\mathbf{x})$ by summing the contributions of all Gaussians:
\begin{equation}
d(\mathbf{x}) = \sum_{p} \sigma_p \exp\!\biggl( -\frac{1}{2} \bigl(\mathbf{x} - \mathbf{x}_p\bigr)^\top \mathbf{\Sigma}_p^{-1} \bigl(\mathbf{x} - \mathbf{x}_p\bigr) \biggr)\,.
\label{eqn:opacity-field}
\end{equation}
This field is discretized onto a 3D grid, and candidate voxels for internal filling are identified where the opacity is below a predefined threshold $\sigma_{\text{th}}$.
For each voxel, we cast rays along the six principal axes to detect transitions from low- to high-opacity regions, marking these voxels for internal filling.
Each identified voxel is then initialized with a pre-defined number of Gaussian primitives with uniform color and opacity. We set each new Gaussian's covariance matrix to be spherical with a randomly assigned scale capped by a maximum value (e.g., $10^{-3}$ times the object's overall size).

\subsection{Conditioned Cross-section Inpainting}
\label{sec:cross-section-inpaint}
\begin{figure}[h]
    \centering
\includegraphics[width=0.8\columnwidth]{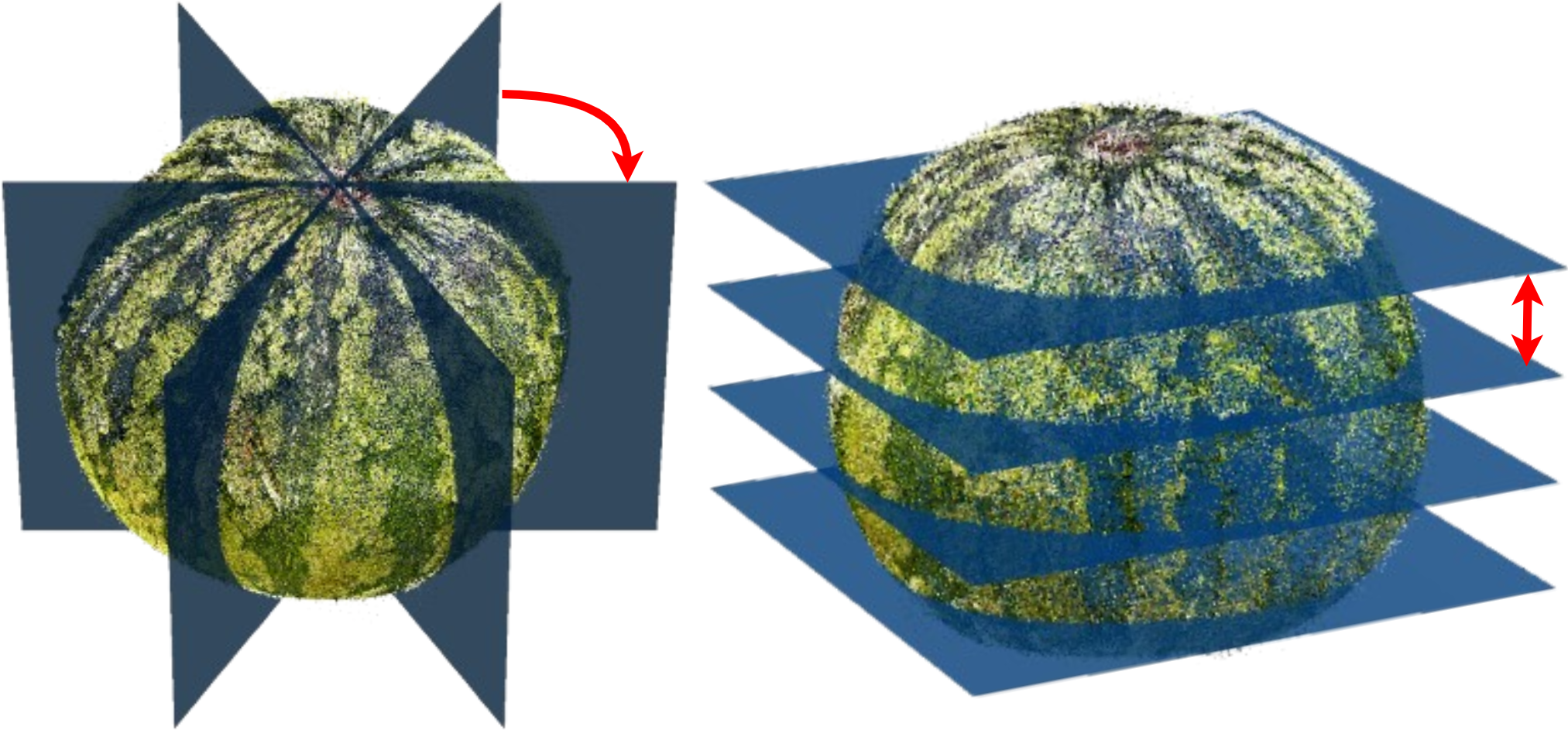}
    \caption{Example of user-defined cross-sections: Left: Vertical cross-sections through the watermelon center, spaced by a specified rotational angle. Right: Horizontal cross-sections, evenly spaced, slicing the watermelon at incremental depths. Note that these input cutting angles are required only during training; once trained, the 3DGS model can be sliced and rendered at arbitrary angles without additional optimization.}
    \label{fig:user_defined_cross}
\end{figure}

As discussed in Section \ref{sec:intro}, obtaining comprehensive cross-sectional data across arbitrary cutting angles is challenging. Fortunately, cross-sectional images of objects from canonical cut angles is easier to acquire. For example, people commonly slice an orange horizontally or vertically, making these views more accessible. Our approach leverages these canonical cross-sectional views to generate the complete internal texture efficiently. Specifically, our method relies on a set of user-specified cutting angles, defined by a cutting plane
\begin{equation}
 a x + b y + c z + d = 0
 \label{eq:cutplane}
\end{equation}
where \( x, y, z \) are 3D coordinates, and \( a, b, c, d \) define the plane's orientation and position. 
These user-defined cross-sections are categorized based on the intrinsic characteristics of the 3D object as illustrated in the Figure \ref{fig:user_defined_cross}, which allows us to apply different text conditions for the diffusion prior. We denote the set of user-defined cross-sectional views as \( \{ V_1, V_2, \dots, V_k \} \), where each \( V_i \) corresponds to a specific cutting plane.

For each cut angle $V_i$ defined by Equation \ref{eq:cutplane}, we create a corresponding 3D mask to retain only the Gaussian primitives close to the cutting plane. During rendering, we mask out the exterior and reveal the slice plane. To account for the varying geometry of rendered cross-sections, we first render RGB images and generate estimated depth maps using a pre-trained depth estimator\cite{birkl2023midas}. These depth maps serve as additional conditioning input to a Stable Diffusion model. We observe that directly optimizing 3DGS parameters for rendered $V_i$ using SDS loss is inefficient in the beginning. During early iterations, many internal Gaussians are untrained, resulting in rendered views that lack distinctive features and hinder SDS convergence. To address this issue, we adopt a two-stage optimization process. In the first stage, we independently perform multiple SDS optimizations on each rendered cross-sectional views \( V = \{V_i \mid i = 1, \dots, k\} \).
For each $V_i$, we employ the depth-conditioned Stable Diffusion model guided by cut-angle specific text prompts (e.g., \textit{"the horizontal cross-sectional view of a watermelon"}). This tailored prompting guides the diffusion model to generate images \( I_{\text{label}}^p \) that align more closely with the orientation and characteristics of each cross-section. The SDS loss can be  formulated as:
\begin{equation}
\mathcal{L}_{\text{SDS}} = \mathbb{E}_{t, \epsilon} \left[ w(t) \left\| \epsilon - \epsilon_\theta \left( \mathbf{I}_{\text{label}}^p + \sigma_t \epsilon, t, e, d \right) \right\|^2 \right],
\label{eq:sds_loss}
\end{equation}
where \( \mathbf{I}_{\text{label}}^p \) is the current reference image at viewpoint \( p \), \( \epsilon \sim \mathcal{N}(0, \mathbf{I}) \) is Gaussian noise added at timestep \( t \), \( \sigma_t \) is the noise level corresponding to timestep \( t \), \( \epsilon_\theta \) is the noise predicted by the diffusion model parameterized by \( \theta \), \( e \) is the text embedding of the cut-angle specific prompt, \( d \) is the depth map corresponding to the cross-sectional view, and \( w(t) \) is a weighting function that balances the contributions from different timesteps.

In the second stage, we use these optimized cross-sectional reference images to update the 3DGS parameters by minimizing the reconstruction loss:

\begin{equation}
\mathcal{L}_{\text{recon}} = \alpha \, \mathcal{L}_{\text{MSE}} + (1-\alpha) \, \mathcal{L}_{\text{SSIM}},
\end{equation}

where \(\alpha\) is a weighting factor, \( \mathcal{L}_{\text{MSE}} \) is the Mean Squared Error between the rendered images \( I_{\text{RGB}}^p \) and the reference images \( I_{\text{REF}}^p \) and \( \mathcal{L}_{\text{SSIM}} \) is the Structural Similarity loss \cite{ssimloss} computed between the same images. This combined loss leverages both pixel-wise differences and perceptual similarities. To preserve the object's external appearance during optimization on cross-sectional views, we also randomly select 10--20 surface views rendered from initial input 3DGS and jointly train them with $V$.

\begin{figure}[h]
    \centering
\includegraphics[width=\columnwidth]{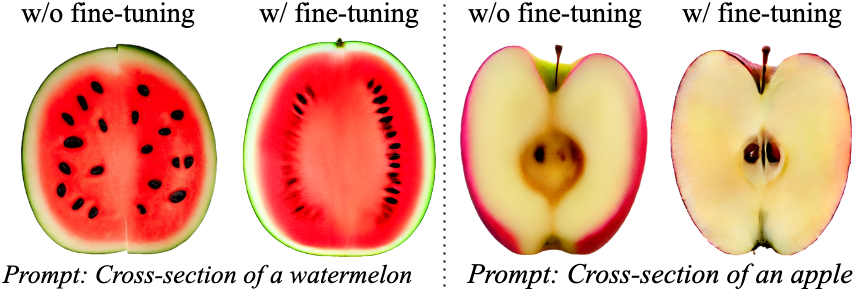}
    \caption{Images generated by stable-diffusion-2-depth\cite{Rombach_2022_CVPR}}
    \label{fig:finetune1}
\end{figure}

\textbf{Fine-tuning with DreamBooth} We noticed that Stable Diffusion models often fail to generate high-quality cross-sectional images as shown in Figure \ref{fig:finetune1}, likely due to the scarcity of such images in the training data. Therefore, we optionally fine-tune the diffusion model using a small set (1--6) of cross-sectional images and the DreamBooth method~\cite{ruiz2023dreambooth}. Each image is paired with an angle-specific text prompt (e.g., \textit{"A vertical cross-section of an object"}) to guide generation. As in DreamBooth, we apply a class-specific prior preservation loss to encourage diverse generation within each category.

\subsection{Progressive Texture Refinement}
\label{sec:refinement}
\begin{figure}[h]
    \centering
\includegraphics[width=0.86\columnwidth]{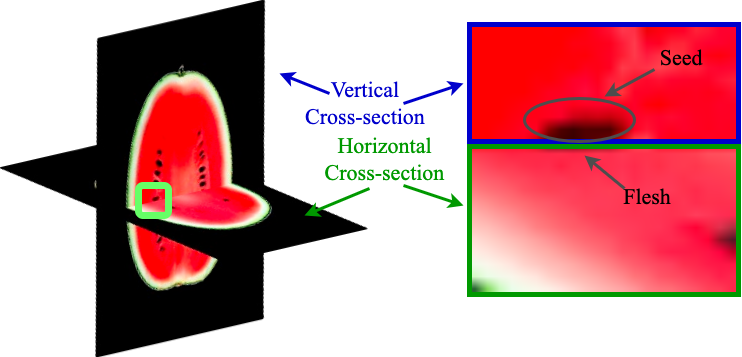}
    \caption{Within the intersected 3D region, the vertical cross-section (top) highlights a seed, while the horizontal cross-section (bottom) depicts the surrounding flesh, which introduces conflicting training signals.}
    \label{fig:consistency1}
\end{figure}

Training 3DGS with reference cross-sectional views (see Section~\ref{sec:cross-section-inpaint}) may introduce spatial inconsistencies. For example, as shown in Figure~\ref{fig:consistency1}, a vertical slice of a watermelon model might display a black seed at a specific location, while a horizontal slice through the same region shows only red flesh.
To mitigate this, we employ an iterative refinement process that jointly optimizes rendering and generation, similar to the fine-tuning stage in DreamGaussians \cite{tang2023dreamgaussian}.
 After each iteration, we render the current cross-sectional views  \( V = \{V_i \mid i = 1, \dots, k\} \) from the trained 3DGS model, then we apply a few additional optimization steps for each reference cross-sectional view using SDS (Equation \ref{eq:sds_loss}). 
We repeat this process until the reconstruction losses for all slices converge to below a predefined threshold $\epsilon$. 
This iterative refinement ensures spatial consistency across the object's cross-sections by harmonizing textures from the trained 3DGS with features from the diffusion model. The algorithm converges when all slices are in agreement, effectively resolving inconsistencies introduced by independently optimized 2D references. 

\textbf{Voxel Smoothing} Due to the discrete nature of input cross-sectional slices $V$, some Gaussian primitives are not covered by the generated masks in Section~\ref{sec:cross-section-inpaint}, leaving them untrained and potentially reducing visual fidelity when they are exposed. To avoid further optimization during rendering, we construct voxel grid over 3DGS and perform smoothing operation for each voxel cell during a predefined interval (e.g., 30--40 iterations). Specifically, untrained Gaussians are assigned colors using a distance-weighted average of nearby trained Gaussians: 
\begin{equation}
 C = \frac{\sum_{i} w_i \cdot C_i}{\sum_{i} w_i}
\end{equation} Here, \( C_i \) represents the color of each nearby trained Gaussian, and \( w_i \) is the inverse distance weight based on the Euclidean distance \( d_i \) between the untrained Gaussian and each trained Gaussian within the same voxel. The grid resolution is chosen empirically to maintain visual quality, ensuring that each voxel contains fewer than 1\% of the total particles. This approach preserves fine texture details while promoting color consistency throughout the model.

\begin{figure*}[t]
    \centering
\includegraphics[width=\textwidth]{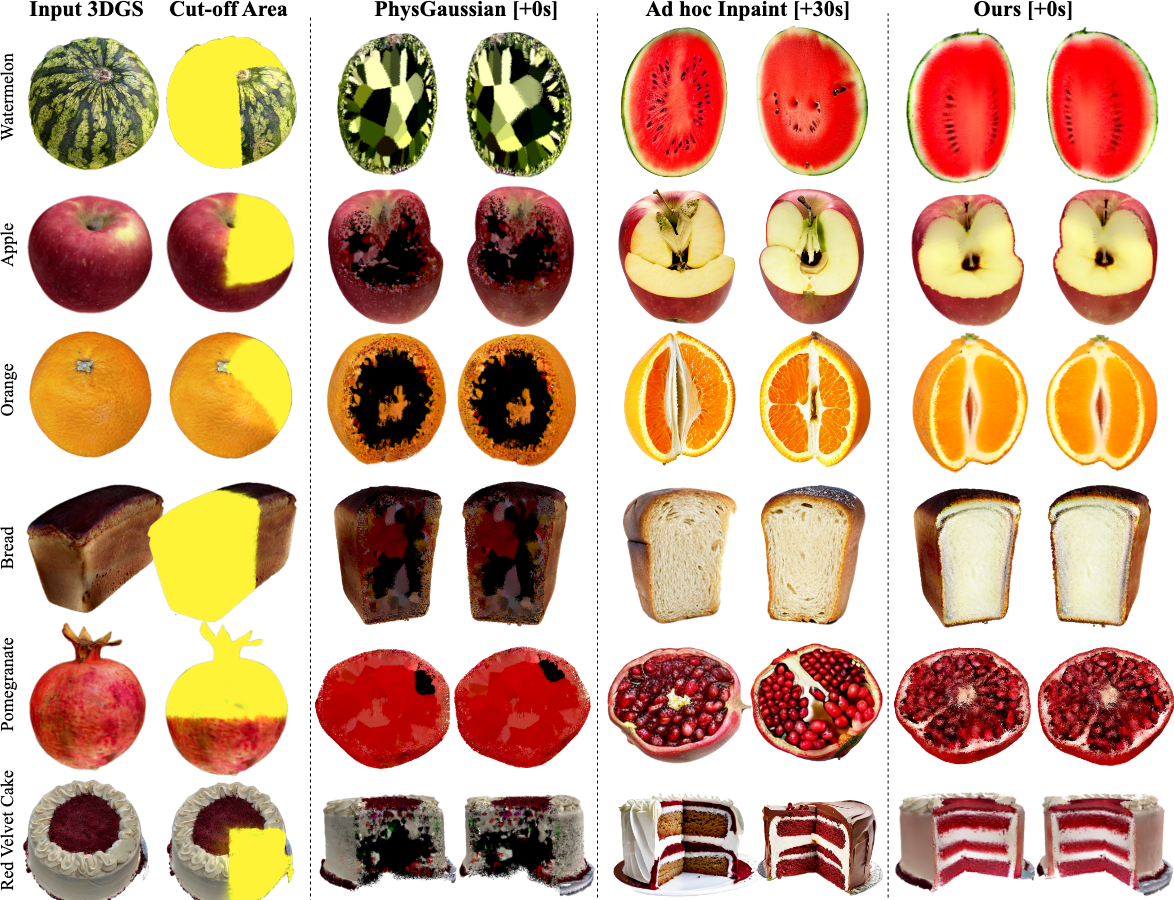}
    \caption{Qualitative comparison with PhysGaussian and 2D inpainting using stable-diffusion-2-depth\cite{Rombach_2022_CVPR} (which requires $\sim$30s for 70 DDIM sampling steps per view). 3DGS from FruitNinja shows better texture consistency and visual quality without any optimization.}
    \label{fig:extensive_res}
\end{figure*}

\subsection{Opaque Atomic Gaussian Particles}
\label{sec:opaque_atom}

To ensure stable training and detailed texture representations, we employ the OpaqueAtom configuration, which incorporates two critical constraints: atomic clipping of Gaussian sizes and uniform high opacity across all Gaussians.

\textbf{Atomic Clipping} 3DGS often optimizes for larger Gaussian primitives during training. However, when these Gaussian particles grow excessively large, they may overlap multiple cross-sectional regions of the 3D object. Since cross-sectional views typically capture fine details in the object's texture, this overlap forces 3DGS to reconcile conflicting representations from various perspectives. Consequently, the training process becomes inefficient and may fail to converge. Besides, large Gaussians lack the spatial resolution needed to represent fine-grained textures accurately and limit precise geometry edits (e.g., slicing a single large Gaussian is not feasible). To counter these, we constrain Gaussian sizes following the approach inspired by AtomGS~\cite{liu2024atomgs} during training. Specifically, we cap each Gaussian’s scale at a minimal fraction (e.g., \(1/3{,}000\)) of the object's dimensions. This fine-grained control preserves texture details for stable training and allows more precise geometric edits.

\textbf{Uniform Opacification}
3DGS renders images by blending contributions from all Gaussians along each ray, including those behind the surface (see Section~\ref{sec:gaussian-rendering}). While this approach efficiently models static 3D objects, it is not reliable in interactive scenarios where geometric changes or user edits alter the ordering and visibility of Gaussian particles dynamically. Particles in the back can unintentionally influence the foreground, leading to unrealistic color distributions. Thus, we assign full opacity to all Gaussian particles to maximize the contribution of front-most Gaussians to the rendered image. This adjustment accurately models opaque materials and ensures that each Gaussian particle faithfully reflects the real-world appearance of the material it represents during geometric deformation.

\section{Experiment}

\subsection{Implementation Details}
\label{sec:implementation_dataset}
We build upon the original 3DGS implementation. For each object, we define one or two types of cross-sections (primarily horizontal and vertical) as shown in Figure~\ref{fig:user_defined_cross}. Horizontal slices are evenly spaced vertically, and vertical slices are arranged radially around the central axis at equal angular intervals. Each training iteration randomly selects 20 surface appearance views rendered from the original 3DGS reconstruction. We use a \(512 \times 512 \times 512\) 3D grid for voxel smoothing. For each reference view, we initially apply 20 SDS optimization steps for generation, followed by 3--4 refinement steps per iteration, as detailed in Section~\ref{sec:refinement}. Training typically required between 120 and 200 iterations. Optionally, we fine-tuned the stable diffusion model \cite{Rombach_2022_CVPR} using 1 to 4 collected cross-sectional views with DreamBooth\cite{ruiz2023dreambooth}, as specified in Section \ref{sec:cross-section-inpaint}.

\textbf{Dataset} To evaluate our method, we collected a dataset comprising six common objects with internal textures distinct from their surface appearances. The dataset includes a watermelon, apple, orange, red velvet cake, loaf bread, and pomegranate. For each object, we captured 160–200 surface images from various angles for initial 3D reconstruction via 3DGS and 1-–4 cross-sectional images (horizontal and/or vertical) online which can be used for fine-tuning (as described in Section \ref{sec:cross-section-inpaint}). We plan to release all datasets and code upon publication.

\begin{figure}[ht]
    \centering
\includegraphics[width=\columnwidth]{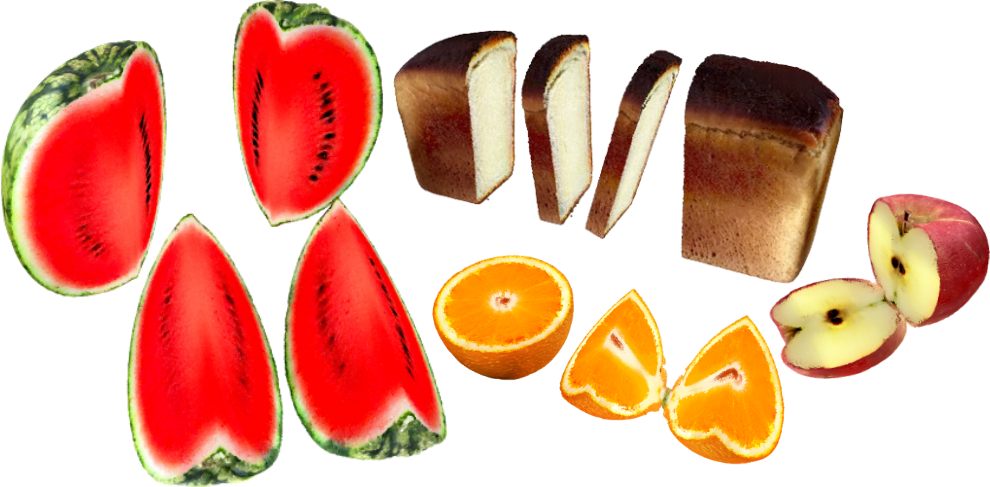}
    \caption{Extensive results. Each item is cut into multiple pieces arbitrarily, showcasing how the cross-sectional textures align consistently across slices.}
    \label{fig:extensive_res}
\end{figure}

\subsection{Qualitative Comparisons}
To the best of our knowledge, no prior work has specifically addressed the inpainting of an object's internal texture for interactive 3D editing use cases. Therefore, we compare our method to two baselines: (1) the internal filling logic from PhysGaussian and (2) a 2D inpainting approach using a depth-conditioned Stable Diffusion model~\cite{Rombach_2022_CVPR}, akin to the ad-hoc inpainting methods employed in VR-GS~\cite{vrgs} and Infusion~\cite{liu2024infusion}. Specifically, we apply identical customized cuts to objects trained using both our method and the original PhysGaussian internal filling logic, then evaluate the exposed internal textures from two distinct camera angles. For the 2D inpainting baseline, we render views from these same camera angles without internal filling and inpaint the textures directly using the Stable Diffusion model with relevant text prompts, such as \textit{'orange, partially cut showing internal structure and flesh'}.

As illustrated in Figure \ref{tab:quantitative_comparison}, the first column presents the input 3DGS, while the second column highlights the cut-off sections of each object. The results indicate that PhysGaussian's generated textures lack realism, appearing overly blurred and unnatural. Directly applying inpainting also struggles to produce faithful views for edited geometry structures, as aligning textures accurately with complex, user-edited shapes is challenging. For example, in the case of the apple, capturing the intricate details of the core, seeds, and uneven inner surface is difficult, especially with cuts made at arbitrary angles or depths. 2D inpainting can also lead to visual inconsistency, as seen with objects like a red velvet cake, where different angles reveal varying textures when inpainted. By comparison, our method generates textures that are highly aligned with the customized cut angles and preserve high visual fidelity. Additionally, 3D models generated by FruitNinja can be rendered in real-time while the ad-hoc inpainting method requires 70 diffusion sampling steps ($\sim$30s), which limits their use in interactive applications. 

Figure \ref{fig:extensive_res} presents extensive results across additional customized geometry modifications, showing that our generated textures consistently maintain high visual fidelity under various cut angles.

\subsection{Quantitative Evaluation}

\begin{table}[h]
    \centering
    \caption{Higher CLIP scores indicate better semantic alignment, while lower KID and FID scores signify superior texture fidelity.}
    \label{tab:quantitative_comparison}
    \setlength{\tabcolsep}{0.1pt}
    \begin{tabular}{lccc}
        \toprule
        \textbf{Method} & \textbf{CLIP Score $\uparrow$} & \textbf{FID $\downarrow$} & \textbf{KID $\downarrow$} x\textbf{$10^{-3}$} \\
        \midrule
        PhysGaussian    & 24.6               & 520.1        & 816.4      \\
        2D Inpainting (Fine-tuned)   & 32.3               & \textbf{176.2}        & \textbf{224.5}     \\
        2D Inpainting   & 25.1               & 314.2        & 536.3     \\
        \textbf{Ours}   & \textbf{33.1}      & 209.2 & 323.7 \\
        \bottomrule
    \end{tabular}
\end{table}

Table~\ref{tab:quantitative_comparison} presents a quantitative comparison of our proposed method against PhysGaussian and a 2D ad-hoc inpainting approach, evaluated on only canonical cross-sections of 3D objects (e.g., horizontal and vertical slices for watermelon). For horizontal cross-sections, we apply random cuts aligned with the chosen orientation to extract the corresponding views. Vertical slices are arranged radially around the central axis at random angles (similar as Figure \ref{fig:user_defined_cross}). We report the CLIP score\cite{hessel2021clipscore} as evaluation metrics, which measures the compatibility between image-caption pairs using category-specific prompts, such as '\textit{the vertical cross-section of a watermelon}'.
Additionally, we use the collected real-world canonical cross-sections images mentioned in \ref{sec:implementation_dataset} to compute the average KID\cite{Kid} and FID\cite{Fid} scores for each rendered view per object. Our method achieves the highest CLIP scores, demonstrating superior semantic alignment with the intended cross-sectional prompts and more accurate texture generation. Moreover, the KID and FID scores of our approach are approximately 60\% better than those of PhysGaussian and are comparable to the 2D inpainting method (with fine-tuning). 

\begin{table}[h]
    \centering
    \caption{Higher cosine similarity indicates better consistency}
    \label{tab:arbitrary_angle_slicing}
    \setlength{\tabcolsep}{0.1pt}  
    \begin{tabular}{lcc}
        \toprule
        \textbf{Method} & \textbf{CLIP Score $\uparrow$} & \textbf{Cosine Similarity $\uparrow$} \\
        \midrule
        PhysGaussian    & 23.9               & 0.89                        \\
        2D Inpainting   & 27.8               & 0.87                        \\
        \textbf{Ours}   & \textbf{29.1}      & \textbf{0.96}               \\
        \bottomrule
    \end{tabular}
\end{table}

To evaluate the texture consistency, we conduct experiments by slicing 3D objects at 120 arbitrary, random angles. Then we compute the average pairwise cosine similarity of CLIP-encoded image features in Tab \ref{tab:arbitrary_angle_slicing}. Using object-specific prompts, such as `\textit{the cross-section of a $\langle$ object $\rangle$}`, we also report the average CLIP score. Results indicate that textures generated by FruitNinja achieve the highest CLIP scores and cosine similarity, demonstrating superior consistency while maintaining fidelity.

\section{Ablation}

\begin{figure}[h]
    \centering
\includegraphics[width=0.7\columnwidth]{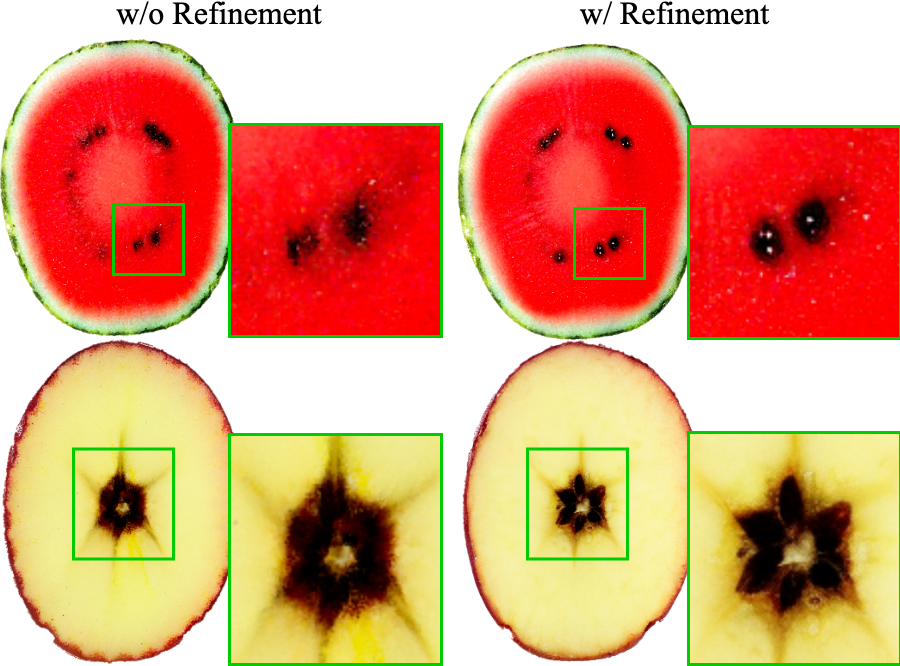}
    \caption{Without continuous texture refinement, reference cross-sectional views often exhibit spatial conflicts as described in Figure \ref{fig:consistency1}, leading to noisy and unrealistic textures in the trained 3DGS models as highlighted in the left column. }
    \label{fig:ablation_refinement}
\end{figure}

\textbf{Progressive Texture Refinement} To assess the necessity of progressive texture refinement (as described in Section \ref{sec:refinement}), we performed an ablation study. Initially, the 3D model was trained using reference 2D views from predefined cross-sectional angles. Texture refinement was then applied ~60 iterations on the two selected objects (a watermelon and an apple). Figure \ref{fig:ablation_refinement} illustrates horizontal cross-sectional views of both objects with (right) and without texture refinement (left). Without refinement, the rendered views exhibit unrealistic textures, characterized by noisy pixels and blurred edges around the seeds, which obscures their shapes and introduces artifacts. These imperfections degrade the visual fidelity of the seeds, making them appear indistinct and unnatural. In contrast, rendered views after refinement show improved visual realism, reduced noise, and enhanced multi-view consistency across cross-sectional perspectives.

\begin{figure}[h]
    \centering
\includegraphics[width=0.82\columnwidth]{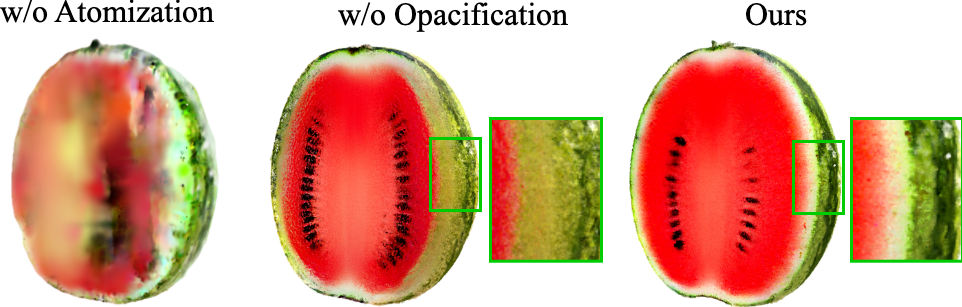}
\caption{
OpaqueAtomGS (right-most) achieves more stable convergence and sharper texture transitions.
}
    \label{fig:ablation_op_atom}
\end{figure}

\textbf{Opaque-Atomic Gaussians} To validate the effectiveness of the OpaqueAtomGS strategy in Section \ref{sec:opaque_atom}, we conducted ablation studies under three configurations: (1) without atomic clipping, (2) without uniform and high opacity, and (3) with full OpaqueAtomGS setting. Each configuration's 3D model was trained for the same number of iterations. As illustrated in Figure \ref{fig:ablation_op_atom}, without atomic clipping, the 3D model struggled to converge and failed to generate realistic textures that align with the references. Without high opacity, the model could not accurately represent the abrupt color transitions at the interface between the white flesh and green surface in the sliced watermelon.

\begin{figure}[h]
    \centering
\includegraphics[width=0.73\columnwidth]{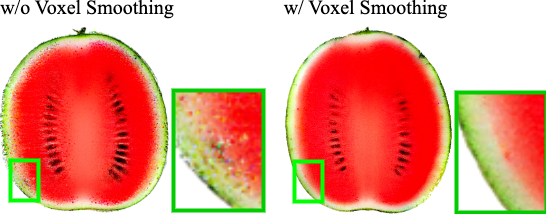}
\caption{Without voxel smoothing (left), Gaussians that are not exposed at the forefront of the cross-sectional plane exhibit random and noisy colors, which reduces overall visual fidelity.
}\label{fig:ablation_voxel}
\end{figure}

\textbf{Voxel Smoothing} Figure \ref{fig:ablation_voxel} shows that voxel smoothing significantly reduces color distortion in untrained Gaussians, enhancing fidelity for internal textures in cross-sectional views outside the training set.

\section{Conclusion}

We present FruitNinja, a novel method for generating realistic internal textures for 3DGS objects, enabling real-time rendering of interior views during unconstrained geometric and topological changes. Experimental results show that FruitNinja significantly improves texture realism and coherence, establishing its effectiveness for immersive 3D applications.

\newpage
{
    \small
    \bibliographystyle{ieeenat_fullname}
    \bibliography{main}
}
\vspace{20pt}
\appendix
\section*{\Large Appendix}

\section{Input Cross-Section Specifications}
\label{sec:x}

In Section \ref{sec:cross-section-inpaint}, we introduced the use of user-defined cross-sections for generating internal textures in Figure \ref{fig:user_defined_cross}. Here, we present the detailed cross-sectional angles used for training the six objects in our experiments. We used two cross-section types for objects in Fig. 11 due to their more complex internal structures, whereas the objects in Fig. 12 used only a single type. Additional visual results for objects after arbitrary geometric transformations demonstrate texture coherence, even when cuts are misaligned with the trained angles.

\begin{figure*}[h]
    \centering
\includegraphics[width=\textwidth]{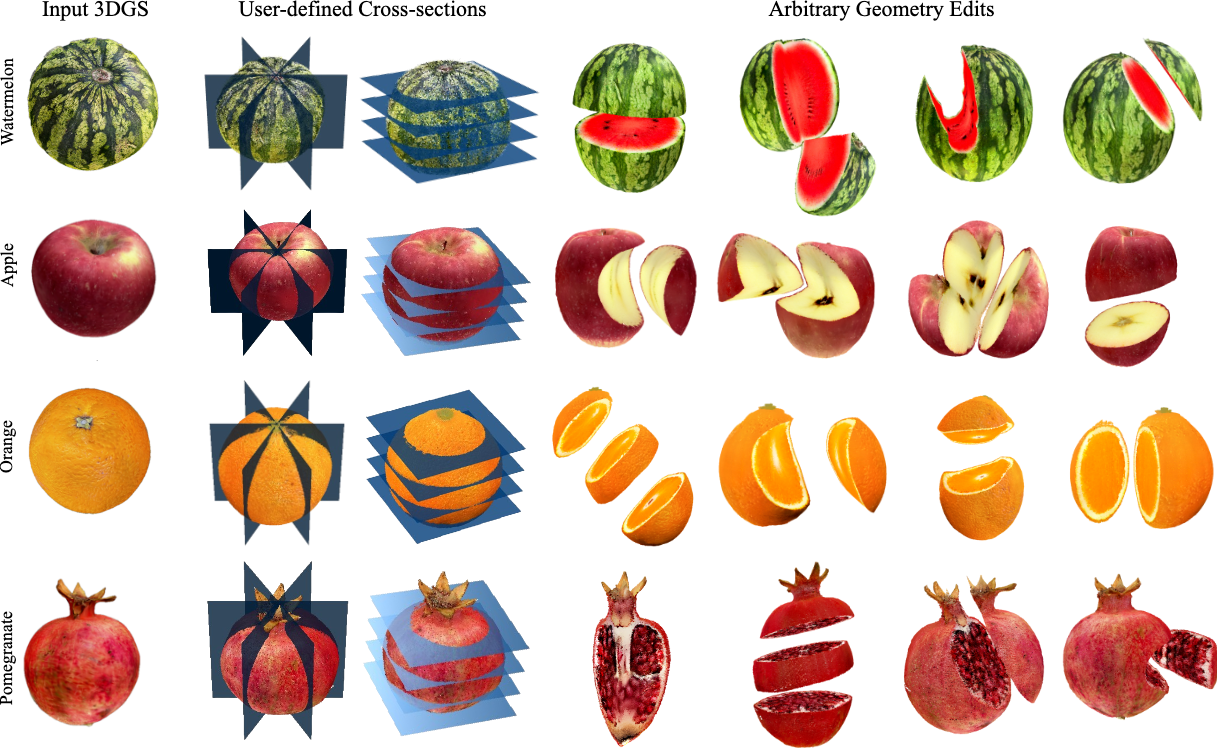}
    \caption{For the four objects shown above (watermelon, apple, orange, and pomegranate), we use two types of input cross-sections: (1) \textbf{Vertical Cross-sections}: 30 slices evenly spaced in angle, spanning a full rotation around the central axis of the object (as illustrated on the second column). (2) \textbf{Horizontal Cross-sections}: We use 40 horizontal slices (as shown in the third column), evenly spaced along the vertical axis to cover the entire object.}
    \label{fig:placeholder}
\end{figure*}

\begin{figure*}[t]
    \centering
\includegraphics[width=\textwidth]{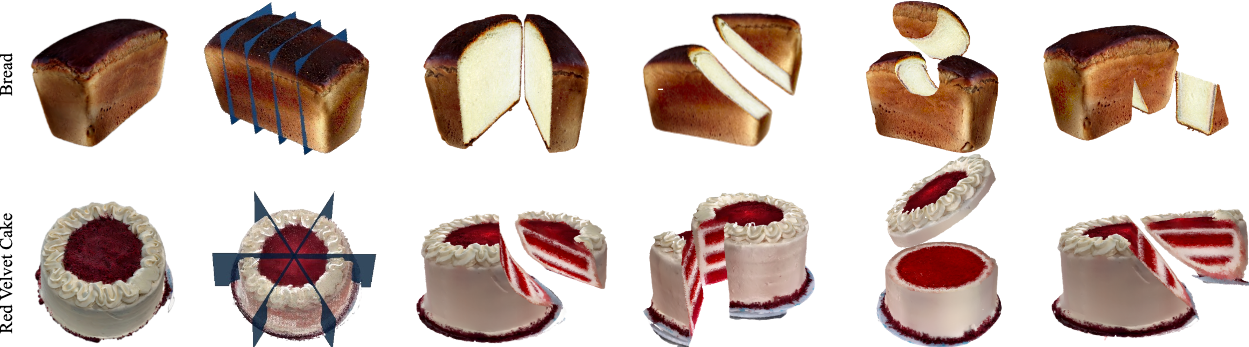}
    \caption{For bread and red velvet cake, we use only vertical cross-sections: (1) \textbf{Bread}: 60 evenly spaced vertical cross-sections (second column) covering the object. (2) \textbf{Red Velvet Cake}: 30 vertical cross-sections, evenly distributed and radially arranged around the central axis, spanning a full rotation of the object.}
    \label{fig:placeholder}
\end{figure*}


\end{document}